\documentclass[letterpaper, 10 pt, conference]{ieeeconf}  

\IEEEoverridecommandlockouts                              
\usepackage[english]{babel}
\usepackage[utf8]{inputenc}	
\usepackage[T1]{fontenc}			
\usepackage{csquotes}
\usepackage{balance}

\usepackage[pdftex]{graphicx}
\usepackage{epstopdf}
\usepackage[svgnames,table,hyperref,usenames,dvipsnames,svgnames]{xcolor}
\usepackage{colortbl}
\usepackage{pgf}
\usepackage{tikz}
\usetikzlibrary{shapes,external}
\usepackage[margin=10pt,indention=.1cm, font=small,labelfont=bf,justification=justified]{caption} 
\usepackage{subcaption}

\usepackage{framed} 
\newenvironment{formal}{
    
    \MakeFramed{\advance\hsize-\width}
    \vspace{2pt}\noindent\hspace{-7pt}\vspace{3pt}
    }{\vspace{3pt}\endMakeFramed}

\let\labelindent\relax
\usepackage{enumitem} 

\usepackage{mathrsfs}	
\usepackage{amsmath, amssymb}
\usepackage[intlimits]{empheq} 
\DeclareMathAlphabet{\mathcal}{OMS}{cmsy}{m}{n}
\allowdisplaybreaks 

\usepackage{amsthm}
\theoremstyle{plain}
\newtheorem{theorem}{Theorem}[section]

\theoremstyle{remark}
\newtheorem{definition}[theorem]{Definition}

\newtheorem{proposition}[theorem]{Proposition}

\usepackage{empheq} 

\usepackage{algorithm}
\usepackage{algorithmic}

\usepackage[sorting=none, style=numeric-comp, giveninits=false, backend=biber,maxbibnames=10,maxcitenames=2,
doi=false,
isbn=false,
url=false,eprint=true]{biblatex}
\AtEveryBibitem{%
  \clearfield{note}%
  \clearfield{pages}
  \clearfield{language}
  \clearfield{number}
  \clearfield{volume}
  \clearfield{url}
  \clearfield{urlyear}
  \clearfield{urlmonth}
}

\renewbibmacro{in:}{}

\usepackage[bookmarks,raiselinks,pageanchor,hyperindex,colorlinks,
citecolor=black,linkcolor=black,urlcolor=black,filecolor=black,menucolor=black
]{hyperref}	
\usepackage[all]{hypcap} 	

\definecolor{genblue}{RGB}{0, 113, 188}
\definecolor{darkblue}{RGB}{27, 20, 100}

\newcommand{\defeq}{\vcentcolon=}

\renewcommand{\.}{
 \ifmmode       
     \cdot 
  \else
    $\,\cdot\,$
 \fi }
 
\newcommand{\+}{
 \ifmmode       
    \, + \,
  \else
    $\,+\,$
 \fi }

\newcommand{\degr}{
 \ifmmode       
    ^{\circ}
  \else
    $^{\circ}$
 \fi }
 
\newcommand{\diff}{\mathrm{d}}

\newcommand{\vektor}[1]{
\begin{pmatrix}
#1
\end{pmatrix}
}

\newcounter{buchstaben}

\setlist[enumerate]{labelindent=*, itemsep=0pt, topsep=1pt, labelwidth=\widthof{2. }, leftmargin=\labelwidth+\labelsep}
\setlist[itemize]{labelindent=*, itemsep=0pt, topsep=1pt}

\newcommand{\mengensymbol}[1]{\ensuremath{\mathbb{#1}}}
\newcommand{\R}{\mengensymbol{R}}

\renewcommand{\^}[1]{
 \ifmmode
    ^\mathrm{#1}
  \else
    $^\mathrm{#1}$%
 \fi }

\makeatletter
\renewcommand\paragraph{%
   \@startsection{paragraph}{4}{0mm}%
      {-.6\baselineskip}%
      {.01\baselineskip}%
      {\normalfont\normalsize\bfseries}}
\makeatother

\graphicspath{{Bilder/}}

\usepackage{xcolor,calc}

\newcommand{\colorNoteBox}[3]{
	\begin{center}
		\vspace{-2ex}\small
		\fcolorbox[rgb]{#1}{#2}{\parbox[t]{\columnwidth-\parindent- 0.2cm}{\setlength{\parskip}{1.5ex}#3}}
	\end{center}
}
\newcommand{\PGnote}[1]{\colorNoteBox{0,1,0}{0.9,0.9,0.9}{PJG: #1}}
\renewcommand{\PGnote}[1]{} 

\title{\LARGE \bf
Learning ODE Models with Qualitative Structure \\Using Gaussian Processes
}

\author{Steffen Ridderbusch,  Christian Offen, Sina Ober-Bl\"obaum, Paul Goulart
\thanks{This work was supported by the EPSRC Centre for Doctoral Training in Autonomous Intelligent Machines \& Systems EP/L015897/1}
\thanks{Steffen Ridderbusch and Paul Goulart are with the Department of Engineering, Control Group,
        University of Oxford, OX1 3PJ Oxford, UK}%
\thanks{Christian Offen and Sina Ober-Bl\"obaum are with the Department of Mathematics, Numerical Mathematics and Control,
        University of Paderborn, 33098 Paderborn, Germany}
\thanks{Correspondence: {\tt\small steffen@robots.ox.ac.uk}}
}%

\begin{document}

\maketitle
\thispagestyle{empty}
\pagestyle{empty}

\begin{abstract}

   Recent advances in learning techniques have enabled the modelling of dynamical systems for scientific and engineering applications directly from data.
   However, in many contexts explicit data collection is expensive and learning algorithms must be data-efficient to be feasible.
   This suggests using additional qualitative information about the system, which is often available from prior experiments or domain knowledge.
   We propose an approach to learning a vector field of differential equations using sparse Gaussian Processes that allows us to combine data and additional structural information, like Lie Group symmetries and fixed points.
   We show that this combination improves extrapolation performance and long-term behaviour significantly, while also reducing the computational cost.

\end{abstract}

\section{Introduction}

Classical parametrized ordinary differential equations (ODEs) are an important tool in science and engineering, as they are a natural and powerful way to model many systems of interest.
These models have been used successfully to understand and predict the behaviour of physical systems, but they can struggle with highly non-linear terms from phenomena that are not well understood, such as friction \cite{naAdaptiveIdentificationControl2018} or chemical reaction dynamics \cite{justiHistoryPhilosophyScience1999, klippensteinTheoreticalReactionDynamics2017}.

To address these shortcomings, the popularity of machine learning has prompted research into learning-based methods for system identification. Many current approaches use state-space Gaussian Process models \cite{kamtheDataEfficientReinforcementLearning2017,deisenrothGaussianProcessesDataEfficient2015,buisson-fenetActivelyLearningGaussian2020}, which learn the flow map of a dynamical system.
The learned function maps a given state to its successor state after a fixed time step, which has the advantage that training data can be directly obtained from trajectory observations.
One difficulty with this approach arises in dynamical systems with complex flows, such as chaotic systems like the Lorenz attractor.  Finding a good flow map in such cases is very difficult, even though the underlying differential equations are comparatively simple. This phenomenon is called \textit{emergence} \cite{newmanEmergenceStrangeAttractors1996}.

In those cases, it is more promising to learn instead the differential equations of the flow.
Rackaukas et al. \cite{rackauckasUniversalDifferentialEquations2020} recently proposed the concept of \textit{Universal Differential Equations} (UDEs), which combine explicitly known terms identified via the SINDy algorithm \cite{bruntonDiscoveringGoverningEquations2016} with neural networks (NNs) to learn the vector field of dynamical systems from trajectories.
They show examples of training a UDE that returns the correct vector field significantly beyond the data initially provided.
However this performance cannot be guaranteed for an unknown system, which is prohibitive in safety-critical applications like control systems.

A popular alternative to NNs that is able to address some of the performance concerns is to use Gaussian Processes (GPs). GPs provide a measure of uncertainty that can be used to judge the accuracy of a given prediction and to identify sections of the state space where more data needs to be gathered.
This general idea has previously been considered by Heinonen et al. \cite{heinonenLearningUnknownODE2018}.
They propose using non-parametric ODEs (npODEs), which consider the values of the vector field at a number of inducing points as additional hyperparameters to be learned such that the resulting field reproduces the original trajectory, which leads to a high-dimensional optimization problem. They focus their results on systems with a stable periodic orbit, but do not explicitly include additional assumptions about the underlying system.

However, most applications do not yield isolated trajectories. Generally some additional qualitative information like long term behaviour or symmetries is available and can be incorporated for data-efficiency.
This concept was recently described as using \textit{side information} by Ahmadi \& Khadir \cite{ahmadiLearningDynamicalSystems2020}. The authors show that it possible to learn a vector field from observations via polynomial regression while also incorporating information like the fixed points or invariant sets as additional constraints.

In this paper we propose a novel combination of learning ODEs with sparse Gaussian processes and structural information that results in learned models with good extrapolation performance. Specifically, we
\begin{itemize}
  \item present a GP model that can describe the vector field of an ODE using sparse GP methods and matrix kernels;
  \item show that this method enables the inclusion of known steady states, resulting in better short-term and long-term predictions;
  \item incorporate Lie Group symmetries at the kernel level, which leads to an accurate preservation of specific first integrals as well as lower computational cost.
\end{itemize}

\section{Background}
\subsection{Problem Setting}
We consider dynamical systems described by ordinary differential equations (ODEs) of the form
\begin{equation}
    \dot{x}(t) = \dfrac{\diff x(t)}{\diff t} = f(x(t)),
    \label{eq:basicode}
\end{equation}
where $x(t)\in\R^D$ is the time dependent state variable and $f: \mathcal{X} \rightarrow \R^D$ is a function defined on a compact set $\mathcal{X} \subset \R^D$, called the \textit{state space}.
We assume $f$ is unknown and are interested in determining it from observations of the solution to the ODE.
For initial value problems, the \textit{trajectory}, or solution, starting from the initial value $x_0$ is defined as a succession of states $x(t)$ via the \textit{flow} $\varphi_t:\R^D\rightarrow\R^D$, as
\begin{equation}
  x(t) = \varphi_t(x_0) = x_0 + \int_0^t f(x(t')) \diff t'.
\end{equation}
We assume that we have $N$ observations of the trajectory
\begin{equation}
    y(t) = x(t) + \epsilon, \qquad \epsilon \sim \mathcal{N}(0, \Omega)
    \label{eq:obs-noise}
\end{equation}
at discrete time points $[ t_1, \ldots, t_N]$.   We further assume that $\epsilon$ is a zero-mean additive  Gaussian noise that is uniform and independent across all dimensions, such that $\Omega = \text{diag}(\sigma_1^2, \ldots, \sigma_D^2)$.
We will assume for simplicity that all components have the same noise variance $\sigma_n^2$.
We are interested in learning the unknown or only partially known function $f$ from the trajectory observations $\mathcal{D}_T = \{ (t_i, y_i)\}_{i = 1,\ldots, N}$ taken from a trajectory of the system of interest in a way that allows us to include additional qualitative information.

\subsection{Structures and Symmetries}
In addition to the general setting above, we consider qualitative information that can be available without knowing the exact form of $f$. This broadens our ability to include additional features into our learned models without collecting more data points.

\paragraph{Fixed points}
For \textit{equilibrium} points or fixed points $\hat{x}$ it holds that ${f(\hat{x}) = \mathbf{0}}$. These points are where the system is at rest and are often known in practice.

\paragraph{Lie group symmetries}
Symmetries in dynamical systems can be used to reduce the computational effort needed to obtain a solution. Further, by Noether's theorem each differentiable symmetry has a corresponding conserved quantity \cite{flasskampSymmetryMotionPrimitives2019} that should be preserved by the learning process.
\PGnote{I don't understand what is meant be 'induce' above.}
In this work we specifically consider Lie group symmetries \cite{hilgertStructureGeometryLie2012}; for further details and motivating examples see the text by Gaterman \cite{gatermannComputerAlgebraMethods2007}.

\begin{definition}[Symmetry Group \cite{flasskampSymmetryMotionPrimitives2019}]
  \label{def:symmetry-group}
  Consider the Euclidean space ${\mathcal{X} = \R^D }$, let $(\mathcal{G}, \circ )$ be a linear Lie Group with representations $\gamma \in \mathcal{L}(\R^D)$, and $\gamma \, x$ the left action of $\mathcal{G}$ on $x\in \mathcal{X}$. 
  We call $(\mathcal{G}, \mathcal{X})$ a \textit{symmetry group} of the system \eqref{eq:basicode} if
  \begin{equation}
    \varphi_t(\gamma\, x_0) = \gamma\, \varphi_t(x_0) \quad \forall (t, \gamma, x_0) \in \R_{\geq0} \times \mathcal{G} \times \mathcal{X}.
    \label{eq:symmetry_def1}
  \end{equation}
\end{definition}
The fundamental linear symmetries in this context are rotations and translations in space \cite{hilgertStructureGeometryLie2012}, but there exist extensions to symmetry groups with non-linear group actions for dynamical systems \cite{russoSymmetriesStabilityControl2011}.


In this work, all considered actions are linear. Therefore, the condition in \eqref{eq:symmetry_def1} can be restated as an equivariance condition on $f$ in the form
\begin{equation}
  f(\gamma \, x) = \gamma \, f(x)  \quad \forall x \in \mathcal{X}.
  \label{eq:equivariance}
\end{equation}
See remark 2 in \cite{flasskampSymmetryMotionPrimitives2019}, for instance.

\paragraph{Second order systems}
Some dynamical systems are most easily modelled as higher order ODEs.
Here, we specifically consider second order systems, like mechanical systems of the form ${\ddot{x}(t) = f(x(t))}$.
They are usually transformed into first order ODEs, but for machine learning purposes it can be both more accurate and more efficient to learn the second order system.

\subsection{Gaussian Process Regression}
\subsubsection{Overview}
For observations ${\mathcal{D} = \{(x_i, y_i)\}_{i = 1,\ldots, N} }$, we define $X \defeq \{x_i\}_{i=1,\ldots,N }$ and $Y \defeq \{y_i\}_{i=1,\ldots,N }$. We further define the collection of unknown noise-free function values as ${F \defeq \{f(x_i)\}_{i=1,\ldots,N } }$.
To learn the underlying function $f$ we use a Gaussian Process (GP), which is a probabilistic non-parametric model \cite{rasmussenGaussianProcessesMachine2006} that assumes that any finite number of function values has a joint Gaussian distribution.
This corresponds to the prior
\begin{equation}
    p(F \,|\, \phi) = \mathcal{N}( F \,|\, \mu(X), K_{X\!,X})
\end{equation}
over the noise-free function values, where ${K_{X\!,X} \in \R^{N\times N}}$ is the covariance matrix with entries $(K_{A\!,B})_{i,j} \defeq k_{\phi}(a_i,b_j)$, $k_{\phi}$ is a chosen covariance function, parametrized by the hyperparameters $\phi$, and $\mu$ is the chosen prior mean.
We assume the observations $Y$ have added Gaussian noise as in \eqref{eq:obs-noise}.
Using Bayes rule and the standard derivation, we find that the GP posterior for an input $x_*$ is
$ p( f(x_*) \,|\, Y, \Omega, \phi) = \mathcal{N}( f(x_*) \,|\, m_*, C_*) $,
with
\begin{align}
  m(x_*) &\defeq \mu(x_*) + K_{x_*, X} ( K_{X\!,X} + \sigma_n^2 I)^{-1} Y \\\label{eq:gpmean}
  C(_*) &\defeq K_{x_*,x_*} - K_{x_*, X} ( K_{X\!,X} + \sigma_n^2 I)^{-1} K_{X, x_*},
\end{align}
where $\Omega$ is the observation noise from \eqref{eq:obs-noise}.


\subsubsection{Differentiating Gaussian Processes}
\label{sec:diffGPs}
To learn the differential equation governing our system, we need data of the form $(x(t), \dot{x}(t))$, which we do not have directly available.
To obtain derivative observations we use the fact that GPs are closed under linear operators.
Using a differentiable kernel $k_{\phi}$ we learn $D$ independent GPs with hyperparameters $\phi_l$ on observations $\mathcal{D}_{T_l} = \{ (t_i, y_l(t_i)\}_{i = 1,\ldots, N}$, one for each component of our initial trajectory.
Following Wenk et al. \cite{wenkODINODEInformedRegression2019}, we obtain the distribution of the time derivatives as
\begin{equation}
  p(\dot{x} \,|\, X, \phi_l) = \mathcal{N}(\dot{x} \,|\, DX, A),
  \label{eq:diffGP}
\end{equation}
where
\begin{align}
  D &= K_{T,T}^{(1)} K_{T,T}^{-1}\\
  A &=  K_{T,T}^{(1,2)} - K_{T,T}^{(1)} K_{T,T}^{-1} K_{T,T}^{(2)}.
\end{align}
Here the superscript $^{(l)}$ indicates the derivative of the kernel with respect to the $l$-th argument, e.g. 
\begin{equation}
  \big(K_{T,T}^{(1)}\big)_{i,j} = \Big( \tfrac{\partial}{\partial a} k_{\phi}(a,b)|_{a=t_i, b = t_j} \Big)_{i,j}
\end{equation}

This allows us to turn data of the form $(t_i, x_i)$ into $(x_i, \dot{x}_i)$.
Given a sufficiently differentiable kernel, we can continue differentiating the GP to obtain higher derivatives.

\subsubsection{Matrix Kernels}
A common setting for GP regression is using scalar kernels $k_{\phi}:\R^D\times\R^D \rightarrow \R$, where multi-dimensional outputs $y\in \R^D$ are addressed with $D$ independent scalar GPs for each dimension \cite{kamtheDataEfficientReinforcementLearning2017,deisenrothGaussianProcessesDataEfficient2015}.
Here, we instead consider the alternative setting of \textit{matrix kernels} $\mathbf{k}_{\phi}: \R^D \times \R^D \rightarrow \R^{D\times D}$. These kernels originate from regularization theory and the work of Aronszajn \cite{micchelliKernelsMultiTask2005}, who showed that they also uniquely characterize a Reproducing Kernel Hilbert Space (RKHS).

In practice, matrix kernels can be constructed from scalar kernels.
One simple example, which corresponds to the matrix kernel used by Heinonen et al. \cite{heinonenLearningUnknownODE2018},  is $\mathbf{k}_{\phi}(x, x') = k_{\phi}(X\!,X') I_D$.
The case of $D$ independent GPs can be covered by using $\mathbf{k}_{\phi} = \text{diag}(k_{\phi_1}, \ldots, k_{\phi_D})$, in which case $\phi \defeq \{\phi_1, \ldots, \phi_D\}$.
Both of these examples fit into the framework of \textit{uncoupled separable matrix kernels} (USM kernels) \cite{wittwarInterpolationUncoupledSeparable2018}, which defines kernels as $\mathbf{k}_{\phi} = \sum k_{\phi_i} Q_i$, where each $Q_i$ is a positive definite matrix.
Using matrix kernels, the covariance matrix $\mathbf{K}_{X\!,X}\in \R^{ND\times ND}$ can be of high dimension, making both training and evaluation computationally costly.
\PGnote{Why is it $\R^{DN\times ND}$ and not $\R^{ND\times ND}$ here?}
However, depending on the structure of the matrices $Q_i$, the covariance $\mathbf{K}$ can have a Kronecker structure, which can be exploited to achieve a reduction in computational cost \cite{wilsonThoughtsMassivelyScalable2015, niatiInverseSumKronecker2019}.
\PGnote{Not sure what you mean by `shape` -- they should all be square and the same size.  Maybe `sparsity`?  I think this bit is not very clear.}

\subsubsection{GIM Kernels}
In order for a learned function to fulfill the equivariance condition \eqref{eq:equivariance}, we consider another class of matrix kernels, the \textit{Group Integration Matrix Kernels} (GIM kernels) introduced by Reisert \& Burkhardt \cite{reisertLearningEquivariantFunctions2007}.
From representation theory, it follows that a chosen kernel $\mathbf{k}_{\phi}$ uniquely characterizes a RKHS $\mathcal{H}$ of the form
\begin{equation}
  \mathcal{H} = \text{span}\{ \mathbf{k}_{\phi}(\cdot, x')y \,|, x'\in \mathcal{X}, y \in \mathcal{Y})\},
\end{equation}
due to the reproducing property, where each element is a function $f:\mathcal{X}\rightarrow \mathcal{Y}$. Therefore, we can construct a Hilbert Space of functions with a desired equivariance by constructing a kernel with that equivariance.
\begin{proposition}[\cite{reisertLearningEquivariantFunctions2007}]
  Let $\mathcal{G}$ be a linear Lie Group with representation and group action as in Def. \ref{def:symmetry-group}, and $k_{\phi}: \mathcal{X}\times \mathcal{X} \rightarrow \R$ be a $\mathcal{G}$-invariant scalar kernel, such that $k_{\phi}(\gamma\, x, \gamma\, x) = k_{\phi}(x,x)$.
  Then
  $\mathbf{k}_{\phi}: \mathcal{X}\times \mathcal{X} \rightarrow \mathcal{L}(\mathcal{Y})$ is a matrix kernel defined by
  \begin{equation}
    \mathbf{k}_{\phi}(x,x') = \int_{\mathcal{G}} k_{\phi}(x,\gamma\, x') \gamma ~\diff \gamma,
    \label{eq:GIMkernel}
  \end{equation}
  where $\int_{\mathcal{G}}\diff \gamma$ is the Haar-integral (see \cite{hilgertStructureGeometryLie2012}), or a sum for finite Lie groups. For $x,x' \in \mathcal{X}$ and $\gamma, \nu \in \mathcal{G}$ it holds that
  \begin{enumerate}
    \item $ \mathbf{k}_{\phi}(\gamma \, x, \nu\, x') = \gamma\, \mathbf{k}_{\phi}(x,x') \nu^{\dag} $
    \item $\mathbf{k}_{\phi}(x,x') = \mathbf{k}_{\phi}(x',x)^{\dag}$
  \end{enumerate}
  Here $\dag$ denotes the conjugate transpose. \hfill $\blacksquare$
\end{proposition}

When using an equivariant kernel it is important to note the implications on the conditioning of the kernel matrix.
Numerical problems due to nearly numerically singular covariance matrices are common in the use of GPs \cite{zimmermannConditionNumberAnomaly2015}, and can result from closely co-located points.
For GIM kernels, we need to consider the orbit $\Gamma(x) = \{\gamma x \,|\, \gamma \in \mathcal{G}\}$ of a given point $x$.
Given an $\mathcal{G}$-equivariant kernel $\mathbf{k}_{\phi}$, it follows that for two points $\hat{X}= \{x, \gamma\,x\}$, where $x \in \R^D, \gamma \in \mathcal{G}$, the covariance matrix $ \mathbf{K}_{\hat{X}, \hat{X}} \in \R^{2D\times 2D} $ is only of rank $D$.

From a probabilistic viewpoint this means that $x$ and $\gamma \, x$ are perfectly correlated, i.e. they are effectively the same sample.
Within the context of group theory it follows that both points are in the same equivalence class and we can reduce the input space $\mathcal{X}$ to the quotient space $\mathcal{Q} = \mathcal{X}/\mathcal{G}$, which has dimension $\text{dim}\,\mathcal{X} - \text{dim}\,\mathcal{G}$ given regularity assumptions on the action.
The dimension of $\mathcal{G}$ can be understood as the dimension of the parameter space that parametrizes the group action representation $\gamma$. For example, the group of planar rotations $\mathcal{G}=SO(2)$ can be parametrized via a rotation angle $\rho$, which means that $\text{dim}\,\mathcal{G} = 1$.

\subsection{Sparse GPs}
\label{sec:sparseGP}
Lastly we address the known issue that training GPs scales as $\mathcal{O}(N^3)$ since   it features the inverse of the covariance matrix.
For dense matrix kernels this problem is even more pressing, as they result in $\mathcal{O}((DN)^3)$ scaling.
Additionally, the known trajectory of the system we are interested in learning may pass some neighbourhood in the state space multiple times, providing redundant data points, especially in the case for periodic systems or near-periodic systems.

One approach for these issues is to use a sparse approximation \cite{quinonero-candelaUnifyingViewSparse2005} that uses $M$ latent outputs $U := \{u_i\}_{i=1,\ldots,M }$ as \textit{inducing variables} on input locations $Z := \{z_i\}_{i=1,\ldots,N }$ with $z_i\in \mathcal{X}$. This reduces the computational complexity to $\mathcal{O}((DM)^3)$.
A simple variant of this is to take a subset of the available observations, but it is also possible to specify arbitrary locations of the state space.
Since we would like to learn the unknown function $f$ across $\mathcal{X}$, we instead consider a regular equidistant grid of inducing points $Z$ covering the region of interest.
For stationary kernels, this results in Block-Toeplitz-Toeplitz-Block (BTTB) structure, for which a small number of specialized algorithms exists \cite{alonsoEfficientParallelAlgorithm2005}.
We will see in the following that, for GIM kernels, it is possible to reduce the computational effort further through specific choices of inducing points.

However, this means that in addition to specifying the inducing inputs $Z$, we must also determine the latent variables $U$. Heinonen et al. \cite{heinonenLearningUnknownODE2018} treat $U$ as another set of hyperparameters to be learned. Their cost function essentially compares the trajectory obtained by integrating over the mean of $p(f(x_*) | Z, U,\phi)$ to the original trajectory.
The downside of this approach is that variance estimation becomes difficult, as the learned latent variables $U$ are assumed to be error-free regardless of their inducing inputs position relative to the original trajectory.

Instead, we use the approach described in {Sec. \ref{sec:diffGPs}} to obtain input-output data of the form $(x, \dot{x})$ to enable better uncertainty quantification over the state space.
It allows us to infer the inducing outputs as described in \cite{quinonero-candelaUnifyingViewSparse2005}. We further choose the \textit{Fully Independent Training Conditional (FITC)} approximation introduced by Snelson \& Ghahramani \cite{snelsonSparseGaussianProcesses2006}, which further reduces computational cost by assuming that all function values $F$ are independent. This is a valid assumption as the evolution of a dynamical system is only dependent on the current state.
Therefore we use an approximate distribution $q(Y\,|\, U) \simeq p(Y\,|\, F)$ with
\begin{equation}
  q(Y\,|\, U) = \mathcal{N}\big(K_{X,Z}K_{Z,Z}^{-1}U, \text{diag}(K_{X\!,X}-Q_{X\!,X})+\sigma_n^2 I \big),
\end{equation}
where $Q_{A,B} \defeq K_{A,Z}K_{Z,Z}^{-1}K_{Z,B}$. The predictive distribution is $q(f(x_*)\,|\, Y, \Omega, \phi) = \mathcal{N}(m^q_*, C^q_*)$ with
\begin{align}
  \label{eq:sgpmean}
  m^q_* &\defeq K_{x_*,Z}\Sigma K_{Z,X}\Lambda^{-1}Y \\
  C^q_* &\defeq K_{x_*,x_*} - Q_{x_*,x_*} + K_{x_*,Z} \Sigma K_{Z, x_*},
\end{align}
where $\Sigma = (K_{Z,Z}+K_{Z,X}\Lambda^{-1} K_{X,Z})^{-1}$ dominates the cost with $\mathcal{O}\big((DM)^3\big)$, and $\Lambda = \text{diag}(K_{X\!,X}-Q_{X\!,X})+\sigma_n^2 I$.

As a result, the output at each inducing location has a variance that reflects its position relative to the original trajectory.
While Snelson \& Ghahramani \cite{snelsonSparseGaussianProcesses2006} consider the inducing inputs as hyperparameters, creating a high-dimensional optimization problem, we pre-determine $Z$ both for ease of use and to exploit the structure of the covariance matrix.
In addition to the above, there are other inducing point methods, notably the variational approach by Titsias \cite{titsiasVariationalLearningInducing2009}. Bauer et al. \cite{bauerUnderstandingProbabilisticSparse2017} conclude that it tends to provide a more accurate approximation than FITC but is more likely to get stuck in local minima during optimization. Comparing these approaches is a matter for future research and for this work we restrict ourselves to FITC.

\section{Learning Differential Equations}

To learn the dynamical system from a given trajectory $\mathcal{D}_T = \{(t_i, y_i)\}_{i=1,\ldots,N}$, we first need time derivatives $\dot{x}$ along the trajectory.
While we could use finite difference schemes, we instead fit another GP and use Sec. \ref{sec:diffGPs} to obtain derivative observations.
In this work we use the GP mean for $x$ and $\dot{x}$, and assume only a small noise term on $\dot{x}$ to improve the conditioning of the kernel matrix, but for future work we aim to incorporate the uncertainty from $Y$.

For all GPs in this work we choose the squared exponential kernel
\begin{equation}
  k_{\phi}(x,x') = \lambda^2 \exp\left( -\dfrac{1}{2}\sum_{j=1}^D \dfrac{(x_j-x'_j)^2}{l_j^2} \right),
\end{equation}
where $\lambda^2$ is the signal variance, $l_j$ a length scale and $\phi \defeq (\lambda, l_1, \ldots, l_D)$, and assume a zero prior mean.
It is possible to train one GP per component of the trajectory, but we found that the benefit relative to using the same kernel across all dimensions was not sufficient to justify the effort.
We minimize the standard negative log likelihood
\begin{equation}
  \begin{split}
    -\log p(Y \,|\, \phi) \propto~& Y^T (K_{T,T}+\sigma_n^2 I)^{-1} Y \\
    &+ \log \det (K_{T,T}+\sigma_n^2 I).
  \end{split}
  \label{eq:derivfit}
\end{equation}
From this GP we obtain data ${\mathcal{D} = \{(x_i, \dot{x}_i)\}_{i=1,\ldots,N}}$, where $ x_i = m(t_i) $ and $\dot{x}_i = \dot{m}(t_i)$ (see \eqref{eq:gpmean}).

\begin{algorithm}[t]
  \caption{Learning a vector field}
  \label{alg:Learn}
  \begin{algorithmic}[1]
    \REQUIRE Trajectory samples $\mathcal{D}_T = \{(t_i, y_i)\}_{i=1,\ldots,N}$, \\ matrix kernel $\mathbf{k}_{\phi}$, inducing inputs $Z$
    \STATE \textit{Step 1: Obtain} $(x, \dot{x})$ \textit{observations}
    \STATE Fit $D$ independent scalar GPs to $\mathcal{D}_T$ by minimizing \eqref{eq:derivfit} via Nelder-Mead.
    \STATE Obtain $X = \{x_i\}_{i=1,\ldots,N}$ via mean $m(t_i)$ of trained GP, and $\dot{X} = \{\dot{x}_i\}_{i=1,\ldots,N}$ via $D X$ (see \eqref{eq:diffGP})
    \STATE \textit{Step 2: Sparse GP for vector field}
    \STATE Learn hyperparameters $\phi$ for given $\mathbf{k}_{\phi}$ and $Z$, \\ minimizing \eqref{eq:spGPfit} via Nelder-Mead.
    \STATE Compute and store $\Sigma K_{Z,X}\Lambda^{-1}Y$ for \eqref{eq:sgpmean}
    \RETURN GP-ODE model

  \end{algorithmic}
\end{algorithm}

From here, we learn the unknown system $\dot{x} = f(x)$.
\PGnote{What does 'approach' mean here?}
We select a suitable compact input space $\mathcal{X}$, and choose suitably covering inducing input locations $Z$ and a matrix kernel $\mathbf{k}_{\phi}\in \mathcal{L}(\R^D)$.
To learn the hyperparameters, we minimize the negative log likelihood $p(\dot{X} \,|\, \phi)$  via the following proportional RHS using the definitions for $\Lambda$ and $Q$ from  Sec. \ref{sec:sparseGP}
\begin{equation}
  \begin{split}
    -\log p(\dot{X} \,|\, \phi) \propto~& \dot{X}^T (\mathbf{Q}_{X\!,X}+{\Lambda})^{-1} \dot{X} \\
    &+ \log \det (\mathbf{Q}_{X\!,X}+{\Lambda}).
  \end{split}
  \label{eq:spGPfit}
\end{equation}
\PGnote{Which side of the above is actually being minimized?  The $\propto$ is confusing here.}

As a result, we obtain a non-parametric expression for $\dot{x} = f(x) \defeq m^q(x)$ using \eqref{eq:sgpmean} and the trained $\mathbf{k}_{\phi}$, which we can integrate using a numerical ODE solver.

As we will show in the following examples, this setting lets us add additional qualitative information.
Known fixed points, or any other states for which derivative information is available, can be added directly to the data $\mathcal{D}$.
Lie Group symmetries are incorporated via an appropriate GIM kernel. Lastly, we can obtain estimates for higher derivatives to learn systems of the form $\ddot{x} = f(x)$.
\PGnote{Is the ``$\ddot{x}(t_i) = \ddot{m}(t_i)$`` actually needed here?  What is $m$?}

\begin{figure*}[ht]
  \centering
  \includegraphics[width = 0.9\textwidth]{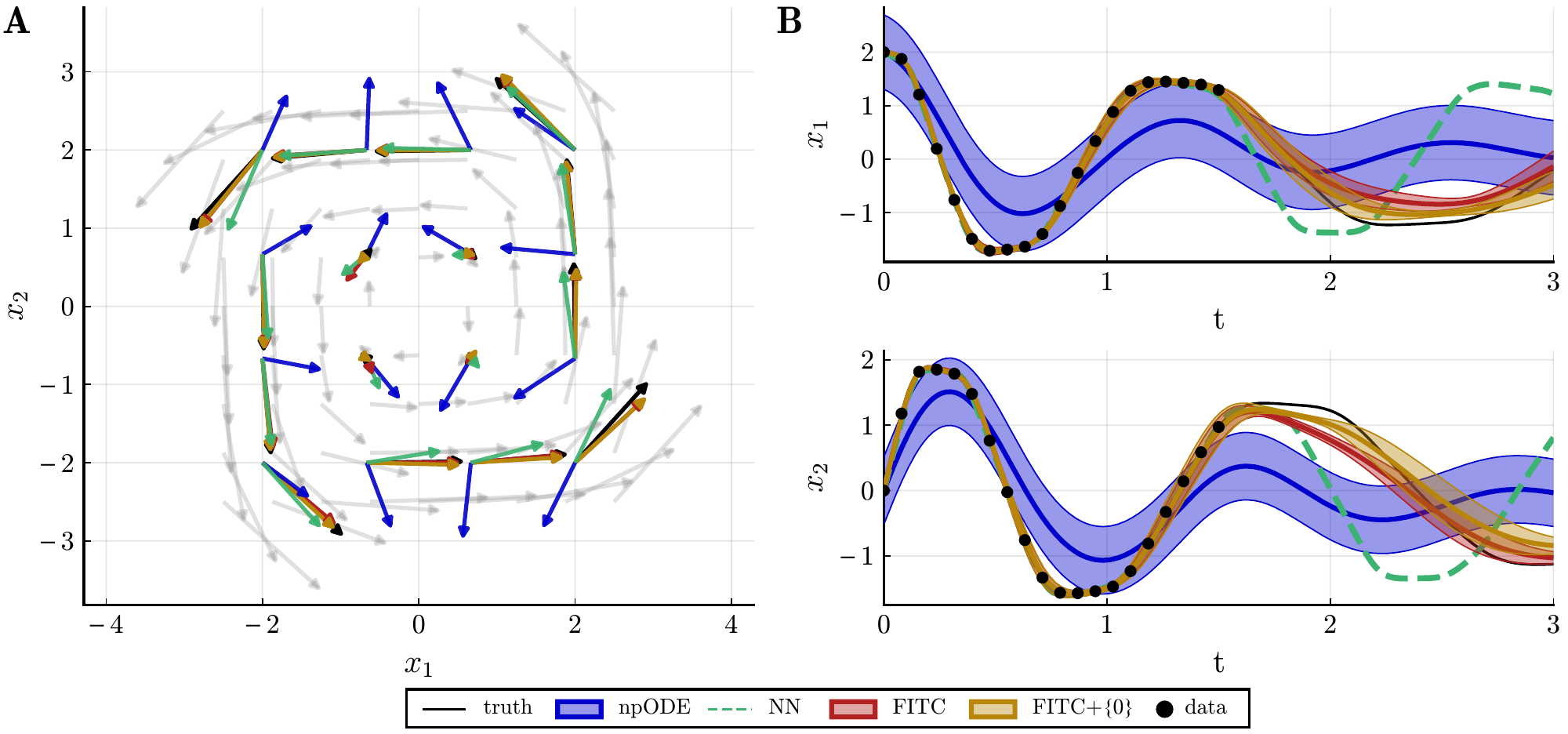}
  \caption{Cubic ODE from Example 4.1 \textbf{A}: Quiver plot of the vector field on the inducing points $Z$, additional points of the real vector field in grey. Each arrow is scaled to $\log10$ of its norm. \textbf{B}: Mean trajectories from the learned
  systems, and pointwise indication of one standard deviation to indicate uncertainty.}
  \label{fig:cubic}
\end{figure*}

\section{Examples}
We next show the effectiveness of our proposed method via two illustrative examples. We consider the example of a simple spiral ODE as well as the Kepler problem \cite{ober-blobaumGalerkinVariationalIntegrators2017}.
\footnote{The code to reproduce the  following results can be found at \texttt{https://github.com/Crown421/StructureGPs-paper}}

\subsection{Spiral ODE}

We first consider the example of an ODE of the form
\begin{equation}
  \dot{x} = \vektor{-0.1 & 2\\ -2 & -0.1} \vektor{x_1^3\\x_2^3},
\end{equation}
which has a single stable fixed point at the origin.
The example is taken from the supplementary documentation of the \textit{DiffEqFlux} package by Rackauckas et al. \cite{rackauckasUniversalDifferentialEquations2020}.
It was used to demonstrate learning an ODE of the form  $f(x) = NN(x^3)$  where $NN$ is a neural network with a two layers, 250 weights, and $\tanh$ as activation function.

As in the original example, we compute 20 equidistant data points in the time span $(0, 1.5)$ from a trajectory starting at the initial value $x_0 = (2, 0)^T$ to use as data $\mathcal{D}_T$.
We further select the input space $\mathcal{X}= [-2,2]^2$ and cover it with a grid $M = 4^2$ equidistant points $Z$.

\begin{table}[b]
  \caption{Mean error between true solution and learned models, split between training time span ($t\in [0, 1.5]$), test/ extrapolation time span ($t\in [1.5,3]$), and total time span. }
  \label{tab:errors}
  \vspace{-0.15cm}
  \centering
  \begin{tabular}{llll}
             & \textbf{Train Error} & \textbf{Test Error} & \textbf{Total Error}\\
  FITC       &  0.0276  & 0.2927 & 0.3198 \\
  FITC+\{0\} &  0.0288 &  0.2015  & 0.2296 \\
  NN         &  0.0215  &  1.3921 & 1.4112 \\
  npODE      &  0.8264  &  1.4025 & 2.1755
\end{tabular}
\end{table}

The results are presented in Fig. \ref{fig:cubic} and in Tab. \ref{tab:errors}. We consider our method for the original 20 data points, as well as for an augmented data set that also contains the fixed point in the origin.
We compare this to the Rackauckas et al. Neural Network approach as well as to the npODEs by Heinonen et al. \cite{heinonenLearningUnknownODE2018}.
For both comparisons we used the original code provided by the authors, and note that for the npODE we changed the Matlab optimization algorithm to 'Trust Region' and increased the weight of the error term capturing the difference between the provided data and the integrated trajectory by a factor of 20.

\begin{figure}[t]
  \centering
  \includegraphics[width = 0.48\textwidth]{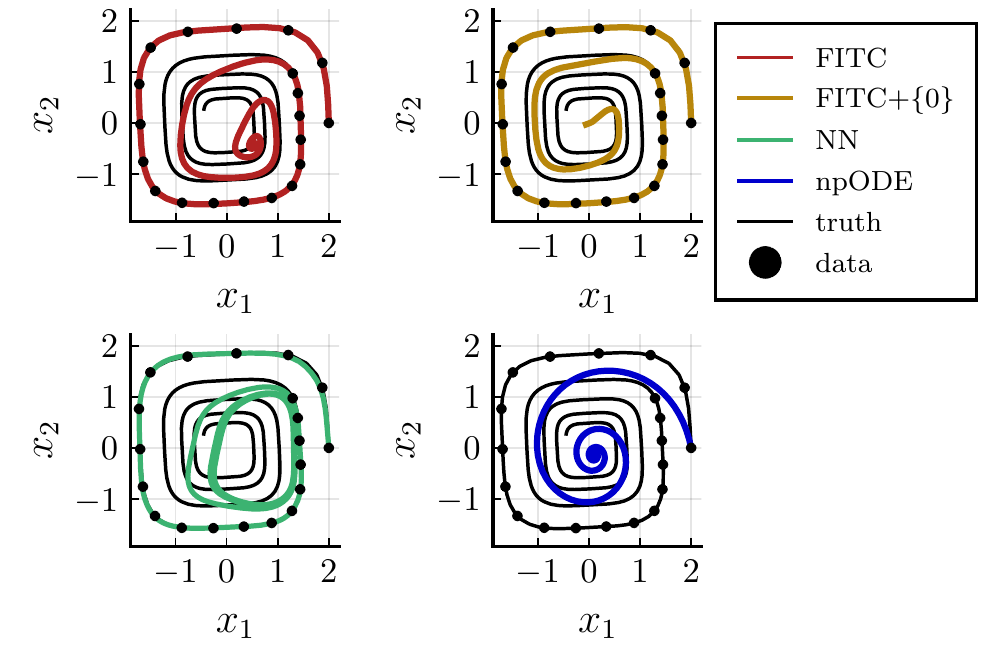}
  \caption{Spiral ODE from Example 4.1. Long-term behaviour of the trajectories integrated over 15 seconds.}
  \label{fig:plot3}
\end{figure}

First we see that the npODE, using the same inducing points $Z$ does not perform well.
While the trajectory has qualitatively similar behaviour to the ground truth, the values of the vector field for the inducing inputs, shown in Fig. \ref{fig:cubic}A, bear little resemblance to the underlying field.
We found that the optimization algorithm got stuck in a local minimum, which could potentially be addressed by using a global optimization method, but is non-trivial due to the high-dimension of their search space.
A second issue is a danger of over-fitting as the trajectory looks better than the inducing outputs would suggest (see also Fig. \ref{fig:plot3}), which may cause low confidence in the accuracy of the vector field even if the integrated trajectory matches the data.

Both the NN approach and ours perform well in the initial range where data is available, but their extrapolation performance differs significantly.
The NN approach diverges quickly when extrapolating, showing an almost periodic orbit in Fig. \ref{fig:plot3}. This could be improved by using a more complex NN, and we note that Rackauckas et al. \cite{rackauckasUniversalDifferentialEquations2020} already include a cubic term as an input to the NN.
Including such transformations into a GP has been considered before, for example in the context of \textit{warped GPs} \cite{snelsonWarpedGaussianProcesses2004}. It suggests interesting further research into additional ways to include known terms into the learning process.

Using only the original data, our approach stays closer to the true trajectory than our comparators.
After appending the fixed point $(x_f, \dot{x}_f) = (\mathbf{0}, \mathbf{0})$ to the data $\mathcal{D}$, the results improve, and in Fig. \ref{fig:cubic}A the (yellow) values on the inducing points match the (black) ground truth values most closely.

We also highlight the advantage of using GPs by including a measure of uncertainty around the trajectory.
For the npODEs we show the sensitivities computed as part of the optimization by the original authors' Matlab code, which are constant for each component.
Since we do not compute the sensitivities, we use a sampling approach to avoid the difficulty of mapping distributions through nonlinear functions. However, we cannot sample the vector field independently but should ideally sample from the distribution over functions $f:\mathcal{X}\rightarrow \R^D$, as noted by Hewing et al. \cite{hewingSimulationTrajectoryPrediction2020}.
We combine their sampling approximation with the work by Bijl et al. \cite{bijlOnlineSparseGaussian2015} and augment the inducing data with up to $100$ past samples and integrate using Euler's method with a small stepsize of $\delta t = 10^{-3}$.
We see that the uncertainty is low for regions where data was available and grows when extrapolating, as intuition would suggest. Further, the ground truth remains mostly within one standard deviation of our predictions.

While the quantitative error increases significantly as the learned system is extrapolated beyond the available data, we consider the qualitative long-term behaviour of the learned system.
As shown in Fig. \ref{fig:plot3}, not explicitly including the fixed point at the origin still leads to a system with an attractor towards which the trajectory converges.
However, the fixed point is substantially off-center and the trajectory distorts increasingly from the shape of the true solution.
Once the fixed point is included, the long-term behaviour improves significantly and the corresponding trajectory converges towards the origin.

This example illustrates that GPs can be very powerful and are able to make accurate short-term predictions beyond the original data, and including known characteristics of the dynamical system under consideration into the model improves the predictive power considerably.
From a dynamical systems perspective, the collected data represents part of the initial transient and the fixed point the long-term steady state.
We see that our approach of learning the vector field allows us to combine both, producing better intermediate predictions.

\subsection{The Kepler System}
Next we consider the 2D Kepler system, which in cartesian coordinates $q\in \R^2$ has the Lagrangian
\begin{equation}
  L(q, \dot{q}) = \dfrac{1}{2}\dot{q}^T\dot{q} + \dfrac{\nu}{\lVert q \rVert},
  \label{eq:Lagrangian}
\end{equation}
where $\nu$ is a positive scalar.
The corresponding second order differential equation of this system is
\begin{equation}
  \ddot{q} = - \dfrac{\nu}{\lVert q \rVert^3} q.
  \label{eq:2ndorderkepler}
\end{equation}
We first consider a first order ODE $\dot{x} = f(x)$ where $x \defeq (q,\dot{q})$. This function is equivariant under the group action represented by
\begin{equation}
  \gamma_{\rho} = \vektor{R_{\rho} & \mathbf{0} \\ \mathbf{0} & R_{\rho}}, \, \text{with} ~ R_{\rho} = \vektor{\cos\rho & -\sin\rho \\ \sin\rho & \cos\rho},
  \label{eq:keplergrpaction}
\end{equation}
which means that the corresponding GIM kernel \eqref{eq:GIMkernel} contains the terms
\begin{subequations}
\begin{align}
  c_{\phi}(x,x') &= \int_0^{2\pi} k_{\phi}(x, \gamma_{\rho}x') \cos\rho \diff \rho  \\
  s_{\phi}(x,x') &= \int_0^{2\pi} k_{\phi}(x, \gamma_{\rho}x') \sin\rho \diff \rho.
\end{align}
\label{eq:keplerkernel}
\end{subequations}
Standard choices for the scalar kernel, like the squared exponential or Mat\'ern kernels, do not admit an analytical solution to these integrals, forcing us to use numerical integration.
The resulting kernel is generally dense but can also represented as an USM kernel, which lets us write the covariance matrix as $\mathbf{K}_{X\!,X} = C_{\phi}(X,X) \otimes Q_1 +  S_{\phi}(X,X) \otimes Q_2$ with appropriate matrices $Q_1, Q_2$.

Given trajectory samples $\mathcal{D}_T = \{(t_i, q_i)\}_{i=1,\ldots,N}$, we fit a standard GP with squared exponential kernel to it and then differentiate twice to obtain data in the form $\mathcal{D}=\{ \big( (q_i,\dot{q}_i), (\dot{q}_i, \ddot{q}_i)\big)\}$.
Further, using the equivariant kernel we must select the location of the inducing inputs $Z$ from the quotient space $\mathcal{Q} = \mathcal{X}/\mathcal{G}$.
The group action \eqref{eq:keplergrpaction} is parametrized by a rotation angle, which means that $\mathcal{Q}$ has dimension $\dim \mathcal{X} -1$.
Hence, we can project the inputs and outputs such that one component is zero without losing any information. We arbitrarily set $q_2 = 0$, for inputs of the form $(\tilde{q}_1, 0, \dot{\tilde{q}}_1, \dot{\tilde{q}}_2)$, and apply the same projection to the input-output pairs in $\mathcal{D}$ to maintain equivariance.
\PGnote{Not sure what is meant by `both' above.   Should it just be dropped?}
As a result, we create a regular grid of inducing points in three dimensions. For scalar kernel $k_{\phi}$ within the GIM kernel we choose the squared exponential kernel when training the GP.

To evaluate the performance of our method on the Kepler system, we consider the following \textit{first integrals}, which correspond to conserved properties.
First, the system preserves the Hamiltonian $H(q, \dot{q}) = \tfrac{1}{2}\dot{q}^T\dot{q} - \tfrac{k}{\lVert q \rVert}$  along the trajectory, which can be understood as an energy-preserving property.
As such, we use a symplectic solver to numerically solve the learned ODE which guarantees good long-term energy behaviour, i.e. no drift in the energy \cite{hairerGeometricNumericalIntegration2006}. In our case we choose the Implicit Midpoint integrator with a small step size of $dt = 0.1$.
Further, the Kepler problem preserves the angular momentum $J$, which means that
$ J(q,\dot{q}) = q_1\cdot \dot{q}_2 - \dot{q}_1 \cdot q_2 $ is invariant along trajectories.

\begin{figure*}[t]
  \centering
  \includegraphics[width = 0.90\textwidth]{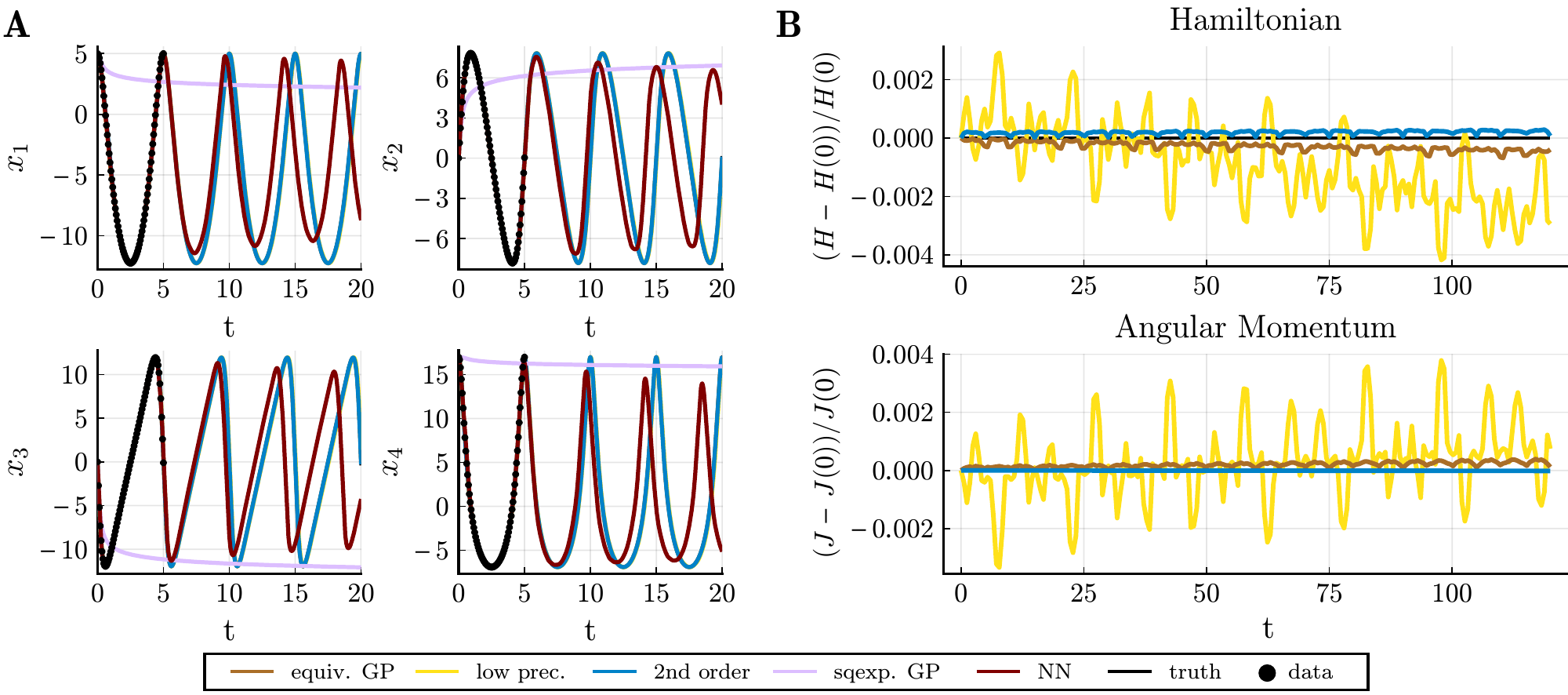}
  \caption{Kepler System. We show the model with the equivariant kernel, the same kernel with fewer quadrature points for integration, the result from directly learning the second order equation \eqref{eq:2ndorderkepler}, and for reference a GP with standard squared exponential kernel as well as a NN based model. \textbf{A:} Trajectories of the learned system. \textbf{B:} The first integrals over extended time.}
  \label{fig:kepler}
\end{figure*}

The first notable result in Fig. \ref{fig:kepler} is that using the squared exponential kernel with $M=1225$ produces unacceptable results. Further, the Neural Network approach, which is similar to the previous example using two layers with $100$ weights each and $\tanh$ as activation function, gives acceptable results, but the learned system is clearly not energy preserving as we see a noticeable decay.

By contrast, using the equivariant kernel with $M=150$ inducing points, we see that all variants overlap very well with the ground truth, to the extent that they are nearly indistinguishable in Fig. \ref{fig:kepler}A. To evaluate them further, we integrate the learned vector fields over a longer time period and consider their first integrals in Fig. \ref{fig:kepler}B.

Here, we did not include the uncertainties in the trajectories, as we need to consider the quotient space for a memory of past samples and also need respect the explicit symmetries, which is subject to ongoing research. 

Despite using a rotation equivariant kernel we note that the angular momentum is not fully preserved, which is attributable to two issues. First there is the additional diagonal noise term in \eqref{eq:sgpmean}, which breaks the equivariance.
While we found that a small amount of noise is necessary to keep the condition number of $\mathbf{K}_{X,X} + \sigma_n^2 I$ tractable, the effect on the equivariance can be reduced by higher signal variance $\lambda^2$ relative to the noise variance $\sigma_n^2$.
This increases the dominant eigenvalues of the covariance matrix by a multiplicative factor, while increasing the smallest eigenvalues sufficiently above zero with the addition of $\sigma_n^2$.
In future work we will consider introducing the noise term in a way that preserves the equivariance of the kernel which will be relevant for learning from noisy measurements as opposed to simulated data points.

Secondly, it is a well known phenomenon that numerical trajectories of Hamiltonian systems are up to exponentially small errors the motions of a modified Hamiltonian system if a symplectic integrator, such as the implicit midpoint rule, is used. 
Therefore, numerical motions exhibit bounded oscillatory energy error behaviour. 
Moreover, symplectic symmetries of the modified system yield modified conserved quantities such that other integrals of motion show oscillatory error behaviour along trajectories as well \cite{hairerGeometricNumericalIntegration2006}. 
As the learnt vector field approximates the Hamiltonian vector field of the Kepler problem, we see such a favourable behaviour in our simulation as well (Fig.~\ref{fig:kepler}B). 
However, a slight drift of the integrals $H$ and $J$ can be spotted as the integrated vector field is not exactly Hamiltonian. 
It is mainly visible in experiments with very low precision.

Lastly, we assume that we know that the dynamical system is second order, and use a GP to learn the second order term in \eqref{eq:2ndorderkepler}.
This function is equivariant under rotation, meaning the group action $\gamma_{\rho} = R_{\rho}$, with $R_{\rho}$ defined in \eqref{eq:keplergrpaction}.
The corresponding GIM kernel again consists of the terms in \eqref{eq:keplerkernel}. Further, we see that the quotient space $\mathcal{X}/\mathcal{G}$, where $\mathcal{X} \in \R^2$, is one-dimensional, and we only select $M = 10$ inducing inputs $Z$ on a line.
We choose inducing points of the form $z_i = (z_{i,1}, 0)$, such that it follows that $s_{\phi}(z_i, z_j)=0$. We also project the inputs $\{ (q_{i,1},  q_{i,2})\}$ in $\mathcal{D}$ to $(\tilde{q}_i, 0)$, transforming the outputs correspondingly.
From here, it follows that $\mathbf{k}_{\phi}(z_i,z_j) = c_{\phi}(z_i,z_j)\cdot I_2$, and $\mathbf{K}_{Z,Z} = C_{\phi}(Z,Z) \otimes I_2 \in \R^{DM\times DM}$.
Exploiting the Kronecker structure allows the training cost for the second order ODE to scale with $\mathcal{O}(M^3)$ instead of  $\mathcal{O}\big((2 M)^3\big)$.

The result is presented in Fig. \ref{fig:kepler}, and shows a nearly constant angular momentum.
The Hamiltonian again displays a slight drift. This shows that while many Hamiltonian systems correspond to second order ODEs, the second order structure itself is not sufficient to learn a Hamiltonian system.

In future work it would be interesting to construct GPs whose realizations are exactly Hamiltonian vector fields with the correct symmetries, and include data from additional trajectories, since a single trajectory provides only a single level set of the Hamiltonian.

\section{Conclusion}
Our results show that using GPs to learn the vector field of an unknown dynamical system provides a powerful and flexible way to incorporate additional knowledge into the learning process.
These structures significantly improve the quality and extrapolation capabilities of learned models.

Our method allows the inclusion of such structures, and can be learned efficiently by optimizing only the kernel hyperparameters which results in a much smaller search space than seen in previous similar approaches.

Our results show the potential of learning vector fields, and to fully realize the advantages of Gaussian Processes we aim to address a more efficient estimation of the uncertainty without sampling.




\printbibliography

\end{document}